\PassOptionsToPackage{hyperindex,breaklinks}{hyperref}
\documentclass[pmlr]{jmlr}
\usepackage{longtable}
\usepackage{booktabs}

\newcommand{\A}{\boldsymbol{A}}
\newcommand{\G}{\boldsymbol{G}}

\newcommand{\I}{\boldsymbol{I}}
\newcommand{\V}{\boldsymbol{V}}
\renewcommand{\S}{\boldsymbol{S}}
\newcommand{\SSigma}{\boldsymbol{\Sigma}}

\newcommand{\e}{\boldsymbol{e}}

\newcommand{\h}{\boldsymbol{h}}
\renewcommand{\u}{\boldsymbol{u}}
\renewcommand{\v}{\boldsymbol{v}}
\newcommand{\w}{\boldsymbol{w}}
\newcommand{\x}{\boldsymbol{x}}
\newcommand{\y}{\boldsymbol{y}}
\newcommand{\z}{\boldsymbol{z}}
\newcommand{\hy}{\widehat{y}}
\newcommand{\regret}{\mathrm{regret}}
\newcommand{\uS}{\|\u\|_{\S_{T}}}

\title[Scale-Invariant Unconstrained Online Learning]{Scale-Invariant Unconstrained Online Learning}

\author{\Name{Wojciech Kot{\l}owski} \Email{wkotlowski@cs.put.poznan.pl} \\
  \addr Pozna{\'n} University of Technology, Poland}

\begin{document}

\maketitle

\begin{abstract}
We consider a variant of online convex optimization in which both the
instances (input vectors) and the comparator (weight
vector) are unconstrained. We exploit a natural scale invariance symmetry in
our unconstrained setting: the predictions of the optimal comparator are
invariant under any linear transformation of the instances. 
Our goal is to 
design online algorithms which also enjoy this property, i.e. are
scale-invariant. We start with the case of coordinate-wise invariance, in
which the individual coordinates (features) can be arbitrarily rescaled. 
We give an algorithm, which achieves essentially optimal regret bound in this
setup, expressed by means of a coordinate-wise scale-invariant norm of the
comparator. We then study general invariance with respect to
arbitrary linear transformations. 
We first give a negative result, showing that
no algorithm can achieve a meaningful bound in terms of scale-invariant norm of
the comparator in the worst case. Next, we compliment this result with a
positive one, providing an algorithm which ``almost'' achieves the desired bound,
incurring only a logarithmic overhead in terms of the norm of the instances.
\end{abstract}

\begin{keywords}
Online learning, online convex optimization, scale invariance, unconstrained online learning, linear classification, regret bound.
\end{keywords}

\section{Introduction}

We consider the following variant of online convex optimization
\citep{PLGbook,ShaiBook,Hazan_OCO}. In trials $t=1,\ldots,T$, the algorithm
receives an instance $\x_t \in \mathbb{R}^d$, on which it predicts $\hy_t =
\x_t^\top \w_t$ by means of a weight vector $\w_t \in \mathbb{R}^{d}$.  Then,
the true label $y_t$ is revealed and the algorithm suffers loss
$\ell(y_t,\hy_t)$, convex in $\hy_t$.  The algorithm's performance is evaluated
by means of \emph{regret}, the difference between the algorithm's cumulative
loss and the cumulative loss of a prediction sequence produced by a fixed
comparator (weight vector) $\u \in \mathbb{R}^d$. The goal of the algorithm is
to minimize its regret for every data sequence $\{(\x_t,y_t)\}_{t=1}^T$ and
every comparator $\u$. This framework includes numerous machine learning
scenarios, such as linear classification (with convex surrogate losses)
and regression. 

Most of the work in online convex optimization assumes 
that the instances and the comparator are constrained to some bounded convex sets,
often known to the algorithm in advance. 
In practice, however, such boundedness assumptions are often unjustified: the learner has little prior knowledge on the potential magnitude of instances, while the prior knowledge on the upper bound of the comparator seems even less realistic.
Therefore, there has been much work recently dedicated to relaxing some of these prior assumptions \citep{StreeterMcMahan2012,Orabona2013,McMahanAbernethy2013,McMahanOrabona2014,OrabonaPal2015ALT,OrabonaPal2016NIPS,Luo_etal2016}. Here, we go a step further, dropping these assumptions entirely and treating the instances, the comparator, as well as comparator's predictions as unconstrained.



In this paper, we exploit a natural \emph{scale invariance} symmetry of the unconstrained setting: if we transform all instances by any invertible linear transformation
$\A$, $\x \mapsto \A \x$, and simultaneously transform the comparator by
the (transposed) inverse of $\A$, $\u \mapsto \A^{-\top} \u$, the predictions, and
hence the comparator's loss will not change. 
This means that the predictions of the \emph{optimal} (loss-minimizing) comparator (if exists) are
invariant under any linear transformation of the instances, so that the scale of the weight vector is only relative to the scale of the instances. 
Our goal is to 
design online algorithms which also enjoy this property, i.e. their predictions are invariant under any rescaling of the instances.

Since in the absence of any constraints, the adversary 
can inflict arbitrarily large regret in just one trial by choosing the instance
an/or comparator sufficiently large,
the regret can only be bounded by a data-dependent function $\Psi(\u,\{\x_t\}_{t=1}^T)$, which can be thought of as a penalty for the adversary for having played with sequence $\{\x_t\}_{t=1}^T$ and the comparator $\u$. We incorporate the scale invariance into this framework 
by working with $\Psi$ which depends on the data and the comparator
only throught the \emph{predictions} of $\u$.
As we will see, designing the online algorithms to have their regret bounded by such $\Psi$ will automatically lead to scale-invariant methods.

We first consider a specific form of scale-invariance, which we call \emph{coordinate-wise invariance}, in which the individual instance coordinates (``features'') can be arbitrarily rescaled (which corresponds to choosing transformation $\A$ which is diagonal). One can think of such rescaling as the change of units in which coordinates are expressed. 
Inspired by the work of \cite{Ross_etal2013UAI}, we choose the penalty function to capture the coordinate-wise invariance in the following decomposable form:

\vspace*{-3mm}

\[
  \Psi(\u,\{\x_t\}_{t=1}^T) = \sum_{i=1}^d f(|u_i| s_{T,i}),
\]

\vspace*{-3mm}

\noindent where $s_{T,i} = \sqrt{\sum_{t=1}^T x_{t,i}^2}$ are ``standard deviations''\footnote{Throughout the paper, the terms ``standard deviation'', ``variance'' and ``covariance matrix'' are used informally (the coordinate values are not shifted by their mean), to capture the scale of the instances.} of individual coordinates (so that $|u_i| s_{T,i}$ measures the scale of $i$-th coordinate relative to comparator's weight) and $f(x) = x\sqrt{\log(1+x^2)}$. This particular choice of $f$ is motivated by a lower bound of \citet{StreeterMcMahan2012}, which indicates that such dependency is the best we can hope for. The main result of Section \ref{sec:coordinatewise_scale_invariance} is a scale-invariant algorithm which achieves this bound up to $O(\log T)$ factor. The algorithm is a first-order method and runs in $O(d)$ time per trial. 
We note that when the Euclidean norms of instances and comparator are bounded by $X$ and $U$, respectively, our bound reduces to Online Gradient Descent bound of $O(UX\sqrt{T})$ \citep{gd} up to a logarithmic factor.

We then turn to a general setup in which the instances can be rescaled by arbitrary linear transformations.
A natural and analytically tractable choice is to parameterize the bound by means of a sum of squared predictions:
%
%
\[
  \Psi(\u,\{\x_t\}_{t=1}^T) = f(\uS),
  \qquad \text{where~~} \uS ~\overset{\mathrm{def}}{=}~ \sqrt{\u^\top \S_T \u} = \sqrt{\sum_{t=1}^T (\x_t^\top \u)^2},
\]
%
%
\noindent and where $\S_T = \sum_t \x_t \x_t^\top$ is the empirical ``covariance'' matrix, 
and $f(x) = x\sqrt{\log(1+x^2)}$ as before.
Our first result is a negative one: any algorithm can be forced by an adversary to have a regret at least $\Omega(\uS\sqrt{T})$ already for $d=2$ dimensional inputs. It turns out that such a bound is meaningless, as a trivial algorithm which always predicts zero has its regret bounded by $O(\uS\sqrt{T})$. 

Is is then the end of the story? While the above result suggests that the adversary has too much power and every algorithm fails in this setting, we show that this view is too pessimistic, complementing the negative result with a positive one. In Section \ref{sec:full_scale_invariance} we derive a scale-invariant algorithm that is capable of \emph{almost} achieving the bound expressed by $\Psi$ above, with only a logarithmic dependence on the norm of the instances. The algorithm is a second-order method and runs in $O(d^2)$ time per trial.
\subsection{Related work}
\label{sec:related}

A standard setup in online convex optimization \citep{PLGbook,ShaiBook,Hazan_OCO} assumes that both the instances\footnote{Most papers assume the bound on the (sub)gradient of loss with respect to $\w$, which translates to bound on the instances,
  as $\nabla_{\w} \ell(y,\x^\top \w) = \partial_{\hy} \ell(y,\hy) \cdot \x$.} and the comparator are constrained to some
bounded convex sets, known to the learner in advance.
A recent series of papers has explored a setting in which the comparator is unconstrained and the learner needs to adapt to an unknown comparator norm \citep{StreeterMcMahan2012,Orabona2013,McMahanAbernethy2013,McMahanOrabona2014,Orabona2014,OrabonaPal2016NIPS}. 
Most of these papers (exception being \citet{Orabona2013}), however, assume that the loss gradients (and thus instances in our setup) are
bounded. Moreover, none of these papers concerns scale-invariance. 

Scale-invariant online algorithms were studied by 
\citet{Ross_etal2013UAI}, who consider a setup similar to our coordinate-wise case. They, however, make a strong assumption that all individual feature constituents of the comparator predictions are bounded: $|u_i x_{t,i}| \leq C$ for all $i=1,\ldots,d$
and $t=1,\ldots,T$, where $C$ is known to the learner.
Their algorithm has a bound which 
depends on $\frac{C}{b_{T,i}}$, $i=1,\ldots,d$ (where $b_{t,i} = \max_{q=1,\ldots,t} |x_{q,i}|$), which is in fact the worst-case upper bound on $u_i$; furthermore their bound also depend as on the ratios
$\frac{b_{T,i}}{x_{t_i,i}}$ ($t_i$ being the first trial in which the $i$-th feature $x_{t_i,i}$ is non-zero), which can be made arbitrarily large in the worst case. 
\citet{Orabona_etal2015} study a similar setup, giving a bound in terms of the quantities $C\frac{|x_{t,i}|}{b_{t,i}}$, and the algorithm still requires to know $C$ to tune its learning rate. 
In contrast, we do not make any assumptions on the predictions of $\u$, and our bound depends on the actual values of $u_i$, solely by means of $\sqrt{u_{i}^2 \sum_t x_{t,i}^2}$, $i=1,\ldots,d$.
\citet{Luo_etal2016} consider a setup similar to our full scale-invariant case, but they require an additional constraint that $|\u^\top \x_t| \leq C$ for all $t$, which we avoid in this work. Finally, \citet{OrabonaPal2015ALT} consider a different notion of invariance, unrelated to this setup.

\section{Problem Setup}

We consider a variant of online convex optimization
 summarized in Figure \ref{fig:online_protocol}. 
In each trial $t=1,\ldots,T$, an instance $\x_t \in \mathbb{R}^d$ is presented to the learner, which produces a weight vector $\w_t \in \mathbb{R}^d$ (possibly depending on $\x_t$) and prediction $\hy_t = \x_t^\top \w_t$. Then, the true label $y_t$ is revealed, and the learner suffers loss $\ell(y_t,\hy_t)$. We assume the loss is convex in its second argument and $L$-Lipschitz (where $L$ is known to the learner), i.e. subderivatives of $\ell$ are bounded, $|\partial_{\hy} \ell(y,\hy)| \leq L$ for all $y,\hy$. Two popular loss functions which fall into this framework (with $L=1$) are logistic loss $\ell(y,\hy) = \log(1+\exp(-y\hy))$, and hinge loss $\ell(y,\hy) = (1-y\hy)_+$. Throughout the rest of the paper, we assume $L=1$ without loss of generality. 

The performance of the learner is evaluated by means of
\emph{regret}:
\[
  \regret_T(\u) = \sum_{t=1}^T \ell(y_t,\x_t^\top \w_t) - \sum_{t=1}^T \ell(y_t, \x_t^\top \u),
\]
where $\u \in \mathbb{R}^d$ is a fixed comparator weight vector, and the dependence on the data sequence has been omitted on the left-hand side as clear from the context. The goal of the learner is to minimize its regret for every data sequence $\{(\x_t,y_t)\}_{t=1}^T$ and every comparator vector $\u$. 

We use the ``gradient trick'' \citep{expgrad,ShaiBook}, which exploits
the convexity of $\ell$ to bound 
$\ell(y_t,\hy'_t) \geq \ell(y_t,\hy_t) + \partial_{\hy_t} \ell(y,\hy_t) (\hy'_t - \hy_t)$ for any subderivative $\partial_{\hy_t} \ell(y_t,\hy_t)$ at $\hy_t$. Using this inequality in each trial with $\hy'_t=\u^\top \w_t$, we get:
\begin{equation}
  \regret_T(\u) \leq \sum_{t=1}^T g_t \x_t^\top (\w_t - \u),
\label{eq:gradient_trick}
\end{equation}
where we denoted the subderivative by $g_t$. Throughout the rest of the paper, we will only be concerned with bounding the right-hand side of (\ref{eq:gradient_trick}), i.e. we will treat the loss to be linear in the prediction, $g_t \hy$, with $|g_t| \leq 1$ (which follows from 1-Lipschitzness of the loss).

\begin{figure}[t]
\begin{center}
\begin{tabular}{|l@{\hspace{10pt}}l|}
\hline
\multicolumn{2}{|l|}{At trial $t=1\dots T$:}\\
&Instance $\x_t \in \mathbb{R}^d$ is revealed to the learner. \\
&Learner predicts with $\hy_t = \x_t^\top \w_t$ for some $\w_t \in \mathbb{R}^d$. \\
&Adversary reveals label $y_{t} \in \mathbb{R}$. \\
&Learner suffers loss $\ell(y_t,\hy_t)$. \\
\hline
\end{tabular}
\end{center}
\caption{Online learning protocol considered in this work.}
\label{fig:online_protocol}
\end{figure}

In this paper, contrary to previous work, we do not impose any constraints on the instances $\x_t$ or the comparator $\u$, neither on the predictions $\x_t^\top \u$.
Since in the absence of any constraints, the adversary 
can inflict arbitrarily large regret in just one trial, the regret can only be bounded by a data dependent function $\Psi(\u,\{\x_t\}_{t=1}^T)$, which we henceforth concisely denote by $\Psi_T(\u)$, dropping the dependence on data as clear from the context. An alternative view, which will turn out to be useful, is to study \emph{penalized regret} $\regret_T(\u) - \Psi_T(\u)$, i.e. the regret offset by $\Psi_T(\u)$, where the latter can now be treated as the penalty for the adversary (a related quantity was called \emph{benchmark} by \citet{McMahanAbernethy2013}).
We will
design online learning algorithms which aim at minimizing the penalized regret,
and this will immediately imply a data-dependent regret bound expressed by $\Psi_T(\u)$.
%

As in the unconstrained setup, the predictions of the optimal comparator are invariant under any linear transformation of the instances, our goal will be to design online learning algorithms which also enjoy this property,
i.e. their predictions do not change under linear transformation of the instances. As we will see, the invariance of learning algorithms will
follow from an appropriate choice of the penalty function.
In Section \ref{sec:coordinatewise_scale_invariance}, we consider algorithms invariant with respect to coordinate-wise transformations. 
We then move to full scale invariance (for arbitrary linear transformations) in Section \ref{sec:full_scale_invariance}.


\section{Coordinate-wise Scale Invariance}
\label{sec:coordinatewise_scale_invariance}

In this section we consider algorithms which are invariant under any rescaling of individual features: if we apply any coordinate-wise transformation $x_{t,i} \mapsto a_i x_{t,i}$ for some $a_i > 0$, $i=1,\ldots,d$, $t=1,\ldots,T$, the predictions of the algorithm should remain the same. Such transformation has a natural interpretation as a change of units in which the instances are measured on each coordinate.
The key element is the right choice of the penalty function $\Psi_T(\u)$, which translates into the desired bound on the regret: the penalty function should be invariant under any feature scaling, offset by the corresponding rescaling of the comparator. Inspired by \cite{Ross_etal2013UAI}, we consider the following function which has such a property:
\[
  \Psi_T(\u) = \sum_{i=1}^d f(|u_i| s_{T,i}),
  \qquad \text{where~~} s_{T,i} = \sqrt{\sum_{t=1}^T x_{t,i}^2}.
\]
Quantities $\{s_{T,i}\}_{i=1}^d$ (``standard deviations'') have a very natural interpretation as measuring the scale of individual features, so that $|u_i| s_{T,i}$ measures the scale relative to the comparator's weight. 
It remains to choose the right form of $f$. To this end, we use a lower bound obtained by \citet{StreeterMcMahan2012} (translated to our setup):
\begin{lemma}[\citealp{StreeterMcMahan2012}, Theorem 7]
  Consider the online protocol in Figure \ref{fig:online_protocol} with $d=1$ and
  $x_{t} = 1$ for all $t$. Choose any learning algorithm which guarantees constant regret against comparator $u=0$. Then for any comparator $u \in \mathbb{R}$, there exists 
  a gradient sequences $\{g_t\}_{t=1}^T$ for which $\sum_t g_t x_{t}(w_t -  u) = \Omega\Big(|u|\sqrt{T} \sqrt{ \log (|u|\sqrt{T})}\Big)$.
\label{lem:lower_bound}
\end{lemma}

Since $x_t = 1$ for all $t$, we have $s_{T,1} = \sqrt{T}$, and the theorem suggests that the best dependence on $|u_i|s_{T,i}$ one can hope for is $f(x) = x \sqrt{\log x}$. This motivates us to study the function of the form \citep{McMahanOrabona2014,OrabonaPal2016NIPS}:
\begin{equation}
  f(x) = x \sqrt{\alpha \log(1+\alpha\beta^2 x^2)},
\label{eq:f_function}
\end{equation}
for some $\alpha, \beta > 0$. This particular choice of parameterization will
simplify the forthcoming analysis.
\citet{Ross_etal2013UAI} have shown that if the learner knew the comparator and standard deviations of each feature in hindsight, the optimal tuning of learning rate would result in a regret bound $\sum_i |u_i| s_{T,i}$ (for $g_t \in \{-1,1\}$). We will show that
without any such prior knowledge, we will be able to essentially (up to $\log(T)$ factor) achieve a bound of $\sum_i f(|u_i|s_{T,i})$, incurring only a logarithmic overhead for not knowing the scale of the instances and the comparator.

Note that the problem decomposes coordinate-wise into $d$ one-dimensional
problems, as:
\[
  \regret_T(\u) - \Psi_T(\u)
  = \sum_{i=1}^d \Big( \underbrace{\sum_t g_t x_{t,i}(w_{t,i} - u_i) - f(|u_i|s_{T,i})}_{\text{penalized regret in 1-dim problem}} \Big).
\]
Thus, it suffices to separately analyze each such one-dimensional problem, and the final bound will be obtained by summing the individual bounds for each coordinate.

\subsection{Motivation}
\label{sec:motivation}
Fix $i \in \{1,\ldots,d\}$ and let us temporarily drop index $i$ for the sake of clarity. Our goal is to design an algorithm which minimizes the one dimensional penalized regret:
\[
  \sum_t g_t x_t (w_t - u) - f(|u| s_T),
\]
where $s_T=\sqrt{\sum_t x_t^2}$
and $f$ is given by \eqref{eq:f_function}. 
If we denote $h_t = - \sum_{q \leq t} x_q g_q$,
we can rewrite the penalized regret by:
\[
  \sum_t g_t x_t w_t + \big(u h_T - f(|u| s_T) \big)
  \leq \sum_t g_t x_t w_t + \sup_{u \geq 0} \big(u |h_T| - f(|u| s_T) \big),
\]
where we observed that the worst-case $u$ will have the same sign as $h_T$. We now use some simple facts on Fenchel duality \citep{BoydVandenberghe04}. Given a function $f \colon X \to \mathbb{R}$, $X \subseteq \mathbb{R}$, its \emph{Fenchel conjugate} is $f^*(\theta) = \sup_{x \in X}\{\theta x - f(x)\}$. If $g(x) = f(ax)$ for some $a > 0$, then $g^*(\theta) = f^*(\theta/a)$. Choosing $X=[0,\infty)$, and $a=s_T$, we get that:\footnote{In the excluded case $s_T=0$ the regret is trivially zero as $x_t=0$ for all $t$.}
\begin{equation}
  \sum_t g_t x_t (w_t - u) - f(|u| s_T) \leq \sum_t g_t x_t w_t + f^*\Big(\frac{|h_T|}{s_T}\Big).
  \label{eq:1-dim_penalized_regret_bound}
\end{equation}
We now use Lemma 18 by \cite{OrabonaPal2016NIPS} (modified to our needs):
\begin{lemma}[\citealp{OrabonaPal2016NIPS}]
Let $f(x) = x \sqrt{\alpha \log (1 + \alpha \beta^2 x^2)}$
for 
$\alpha, \beta > 0$ and $x \geq 0$.
Then
$f^*(\theta) \leq \frac{1}{\beta} e^{\frac{\theta^2}{2\alpha}}$. 
\label{lem:francesco_conjugate}
\end{lemma}
Applying Lemma \ref{lem:francesco_conjugate} to \eqref{eq:1-dim_penalized_regret_bound} results in:
\begin{equation}
 \sum_t g_t x_t (w_t - u) - f(|u| s_T) \leq  \sum_t g_t x_t w_t + \frac{1}{\beta} \exp\Big(\frac{h_T^2}{2\alpha s_T^2}\Big).
 \label{eq:1-dim_penalized_regret_dual}
 \end{equation}
 The main advantage of this bound is the elimination of unknown comparator $u$. We can now design learning algorithm to directly minimize the right-hand side of \eqref{eq:1-dim_penalized_regret_dual} over the worst-case choice of the data. What we derived here is essentially a variant of ``regret-reward duality'' \citep{StreeterMcMahan2012,McMahanOrabona2014,OrabonaPal2016NIPS}.

\subsection{The algorithm}

We now describe an algorithm which aims at minimizing \eqref{eq:1-dim_penalized_regret_dual} for each coordinate $i=1,\ldots,d$. The algorithm maintains the negative past cumulative (linearized) losses:
\[
  h_{t,i} = -\sum_{q \leq t} g_q x_{q,i},
\]
as well as variances $s^2_{t,i}$ for all $i$. At the beginning of trial $t$,
after observing $\x_t$ (and updating $s^2_{t,i}$), the algorithm predicts with weight vector $\w_t$, such that:
\begin{algorithm2e}[t]%
\label{alg:one}
\DontPrintSemicolon
\SetAlgoNoEnd
\SetKwInOut{Parameter}{Parameter}
\SetKwInOut{Initialization}{Initialization}
\Parameter{$\alpha > \frac{9}{8}=1.125$.}
\Initialization{$s^2_i \leftarrow 0, \; h_i \leftarrow 0$, $\quad i=1,\ldots,d$.}
 \For{$t=1,\ldots,T$}{
   Receive $\x_t \in \mathbb{R}^d$\;
   \For{$i=1,\ldots,d$}{
    $s^2_i \leftarrow s^2_i + x_{t,i}^2$\;
    \eIf{$s^2_i > 0$}{
    $\eta_i \leftarrow 
    \frac{1}{\alpha td} \exp\Big(\frac{h_i^2+x_{t,i}^2}{2\alpha s_i^2} \Big)$\;
  $w_i \leftarrow \eta_i \frac{h_i}{s_i^2}$\;}{$w_i \leftarrow 0$\;}
   }
   Predict with $\hy_t = \sum_i w_i x_{t,i}$\;
   Receive $y_t$ and suffer loss $\ell(y_t,\hy_t)$\;
   Compute $g_t = \partial_{\hy_t} \ell(y_t,\hy_t)$\;
   Update $h_i \leftarrow h_i - g_t x_{t,i}$ for all $i=1,\ldots,d$\;
 }
\caption{Coordinate-wise scale invariant algorithm}%
\end{algorithm2e}%
\begin{equation}
  w_{t,i} =  \eta_{t,i} \frac{h_{t-1,i}}{s^2_{t,i}} , \qquad 
  \text{where~~} \eta_{t,i} = \frac{1}{\alpha td} \exp\bigg(\frac{h_{t-1,i}^2+x_{t,i}^2}{2\alpha s_{t,i}^2} \bigg).
\label{eq:weights_coordinatewise}
\end{equation}%
Our algorithm resembles two previously considered methods in
online convex optimization, AdaptiveNomal \citep{McMahanOrabona2014} and  PiSTOL
\citep{Orabona2014}. Similarly to these methods, 
we also use a step size which is exponential in the square
of the gradient (which is actually directly related to the same shape of regret
bound \eqref{eq:f_function} we are aiming for). However, we counterweight the
total gradient by dividing it by the variance $s_{t,i}^2$, whereas
AdaptiveNormal uses to this end the number of trials $t$, while PiSTOL -- sum
of the absolute values, $\sum_{q \leq t} |g_t x_{t,i}|$. Only our choice leads
to a scale invariant algorithm, which is easiest to understand by thinking in
terms of physical units: if we imagine that $i$-th coordinate of instances has
unit $[x_i]$, the term in the exponent in \eqref{eq:weights_coordinatewise} is
unitless, while the weight $w_i$ has unit $1/[x_i]$, so that the prediction
$\hy_t$ also becomes unitless. Thus, rescaling the $i$-th coordinate (or,
equivalently, changing its unit) does not affect the prediction. 
%
Note that our algorithm uses a separate ``learning rate'' $\eta_{t,i}$ for each coordinate,
similarly to methods by \citet{McMahanStreeter2010,adagrad}. The pseudo-code is
presented as Algorithm \ref{alg:one}. 

We now show that the algorithm maintains small penalized regret \eqref{eq:1-dim_penalized_regret_dual}. To simplify notation, define the \emph{potential function}:
\begin{equation*}
  \psi_t(h,s) = \left\{ \begin{array}{ll}
      (td)^{-1} \exp\Big(\frac{h^2}{2\alpha s^2}\Big) & \text{when~~} s > 0, \\
      0 & \text{otherwise}.
\end{array} \right.
\end{equation*}
\begin{lemma}
\label{lem:coordinatewise_main_lemma}
Let $\alpha_0 = \frac{9}{8}$ and define:
$\kappa(\alpha) = \exp\big(\frac{1}{2(\alpha - \alpha_0)}\big)$. In each trial $t=1,\ldots,T$, for all $i =1,\ldots,d$, Algorithm \ref{alg:one} satisfies:
  \[
    g_t w_{t,i} x_{t,i} + \psi_t(h_{t,i},s_{t,i}) ~\leq~ \psi_{t-1}(h_{t-1,i},s_{t-1,i}) + \frac{\kappa(\alpha)}{td}.
  \]
\end{lemma}
The proof is given in Appendix \ref{sec:coordinatewise_main_lemma}. Lemma \ref{lem:coordinatewise_main_lemma} can be though of as a motivation behind the particular form of the weight vector used by the algorithm: the algorithm's predictions are set to keep its loss bounded by the drop of the potential. Note, however, that the algorithm does not play with the negative gradient of the potential (which is how many online learning algorithms can be motivated), as there is additional, necessary, correction of $\exp(x_{t,i}^2/(2\alpha s_{t,i}^2))$ in the weight expression.

Applying Lemma \ref{lem:coordinatewise_main_lemma} to each $t=1,\ldots,T$ and summing over trials gives:
\[
  \sum_t g_t w_{t,i} x_{t,i} + \psi_T(h_{T,i},s_{T,i}) \leq \frac{\kappa(\alpha)}{d}
  (1 + \log T),
\]
where we bound $\sum_{t=1}^T \frac{1}{t} \leq 1 + \log T$. 
Identifying the left-hand side of the above with the right-hand side
of (\ref{eq:1-dim_penalized_regret_dual}) for $\beta= Td$, and following the line of reasoning in Section \ref{sec:motivation}, we obtain the bound on the penalized regret for the $i$-th coordinate:
\[
  \sum_t g_t  x_{t,i} (w_{t,i} - u_{i}) \leq |u_i|s_{T,i}\sqrt{\alpha \log(1+\alpha d^2 T^2 u_i^2 s_{T,i}^2 )} + \frac{\kappa(\alpha)}{d}(1 + \log T).
\]
Summing over $i=1,\ldots,d$ results in the following regret bound for the algorithm:
\begin{theorem}
  For any comparator $\u$ and any sequence of outcomes $\{(\x_t,y_t)\}_{t=1}^T$,
  Algorithm \ref{alg:one} satisfies:
\[
  \regret_T(\u) ~\leq~ \sum_{i=1}^ d |u_i|s_{T,i}\sqrt{\alpha \log(1+\alpha d^2 T^2 u_i^2 s^2_{T,i} )} + \kappa(\alpha)(1 + \log T).
\]
\label{thm:coordinatewise}
\end{theorem}

We finish this section by comparing the obtained bound with a standard bound of Online Gradient Descent $UX\sqrt{T}$ when the instances and the comparator are bounded,
$\|\x_t\| \leq X$, $\|\u\| \leq U$. 
By Cauchy-Schwarz inequality we have:
\[
  \sum_i |u_i| s_{T,i} \leq \sqrt{\sum_i u_i^2} \sqrt{\sum_i s^2_{T,i}}
  \leq U \sqrt{\sum_{i,t} x_{t,i}^2} = U \sqrt{\sum_t \|\x_t\|^2} \leq UX\sqrt{T},
\]
so that our bound is $O(UX\sqrt{T} \log(1+d^2 U^2X^2 T^3))$, incurring only a logarithmic overhead for not knowing the bound on the instances and on the comparator in hindsight. 

\section{Full Scale Invariance}
\label{sec:full_scale_invariance}

In this section we consider algorithms which are invariant under general linear transformations of the form $\x_t \mapsto \A \x_t$ for all $t=1,\ldots,T$. As we will see, imposing such a general symmetry will lead to a second order algorithm, i.e. the algorithm will maintain the full covariance matrix $\S_t = \sum_{q \leq t} \x_t \x_t^\top$.
To incorporate the scale invariance into the problem, we choose the penalty $\Psi_T(\u)$ to depend only on the predictions generated by $\u$.
A natural and analytically tractable choice is to parameterize $\Psi_T(\u)$ by means of a sum of squared predictions:
\[
  \Psi_T(\u) ~=~ f(\uS),
  \qquad \text{where~~} \uS  ~\overset{\mathrm{def}}{=}~ \sqrt{\u^\top \S_T \u}
  = \sqrt{\sum_{t=1}^T (\x_t^\top \u)^2}.
\]
As before, by taking into account the lower bound from Lemma \ref{lem:lower_bound}, we
choose $f(x)$ as defined in \eqref{eq:f_function}, i.e.
$f(x) = x \sqrt{\alpha \log(1+\alpha\beta^2 x^2)}$,
for some $\alpha, \beta > 0$. Our goal is thus to design a scale-invariant algorithm
which maintains small penalized regret:
\begin{align*}
  \regret_T(\u) - \Psi_T(\u) &= \sum_t g_t \x_t^\top \w_t
+ \h_T^\top \u - f(\uS) \\
&\leq\sum_t g_t \x_t^\top \w_t
 +  \sup_{\u} \bigg( \h_T^\top \u - f(\uS) \bigg),
\end{align*}
where we defined $\h_t = - \sum_{q \leq t} g_q \x_q$. We will make use of the following general result, proven in the Appendix \ref{appendix:conjugate}. For any positive semi-definite matrix $\A \in \mathbb{R}^{d \times d}$,
let $\|\u\|_{\A} \overset{\mathrm{def}}{=} \sqrt{\u^\top \A \u}$ denote the semi-norm of $\u \in \mathbb{R}^d$ induced by $\A$. We have:
\begin{lemma}
  For any $f(x) \colon [0,\infty) \to \mathbb{R}$, 
  any positive semi-definite matrix $\A$ and any vector $\y \in \mathrm{range}(\A)$,
  \[
\sup_{\u} \Big\{ \y^\top \u - f\big(\|\u\|_{\A}\big) \Big\} 
= f^*\big(\|\y\|_{\A^{\dagger}}\big)
  \]
  where $f^*(\theta) = \sup_{x \geq 0} x \theta - f(x)$ is the conjugate of $f(x)$
  and 
 $\A^{\dagger}$ denotes the pseudo-inverse of $\A$. In particular,
\[
\sup_{\u} \Big\{ \h_T^\top \u - f\big(\|\u\|_{\S_T}\big) \Big\} 
= f^*\Big(\|\h_T\|_{\S_T^{\dagger}}\Big).
\]
\label{lem:conjugate}
\end{lemma}

Application of Lemma \ref{lem:conjugate} together with
Lemma \ref{lem:francesco_conjugate} gives:
\begin{equation}
  \regret_T(\u) - \Psi_T(\u) \leq \sum_{t=1}^T g_t \x^\top \w_t
  + \underbrace{\frac{1}{\beta} \exp \Big(\frac{1}{2\alpha} \|\h_T\|_{\S^{\dagger}_T} \Big) }_{f^*\big(\|\h_T\|_{\S_T^{\dagger}}\big)}
  \label{eq:full_penalized_regret_dual}
\end{equation}
As before, we have eliminated the unknown comparator from the equation, and we will design the algorithm to directly minimize the right-hand side of \eqref{eq:full_penalized_regret_dual} over the worst-case choice of the data. 

\subsection{Lower bound}

We start with a negative result. It turns out that the full scale invariance setting
is significantly harder than then coordinate-wise one already for $d=2$. We will show that any algorithm will suffer at least $\Omega(\|\u\|_{\S_T}\sqrt{T})$ regret in the worst case, and this bound
has a matching upper bound for a trivial algorithm which predicts $0$ all the time.

\begin{theorem}
  Let $d \geq 2$. For any algorithm, and any nonnegative number $\beta \in \mathbb{R}_+$,
  there exist a sequence of outcomes and a comparator $\u$, such that
  $\|\u\|_{\S_T} = \beta$ and: 
  \[
    \regret_T(\u) \geq \|\u\|_{\S_T} \sqrt{T/2}.
  \]
\label{thm:gramian_lower_bound}
\end{theorem}
On the other hand, consider an algorithm which predicts $0$ all the time. In this
case,
\[
  \regret_T(\u) = - \sum_{t=1}^T g_t \x_t^\top \u
  \leq \sum_{t=1}^T |g_t \x_t^\top \u|
  \leq \sqrt{\sum_t (\x_t^\top \u)^2} \sqrt{\sum_t g^2_t}  
  \leq \|\u\|_{\S_T} \sqrt{T},
\]
where the second inequality is from Cauchy-Schwarz inequality.
Thus, the lower bound is trivially achieved, and we conclude that it is not
possible to obtain meaningful bound that only depends on $\|\u\|_{\S_T}$ by any online algorithm in the worst-case.

\subsection{The algorithm}

While it is not possible to get a meaningful bound in terms of $\uS$
in the worst case, here we provide a scale-invariant algorithm which \emph{almost} achieves
that. Precisely, we derive an algorithm with a regret bound expressed by
$f(\uS)$, with $f$ defined as in \eqref{eq:f_function}, with
only a logarithmic dependence on the size of the instances hidden in constant $\beta$.

The algorithm is designed in order to minimize the right-hand side of \eqref{eq:full_penalized_regret_dual}. It maintains the negative past cumulative (linearized) loss vector $\h_t = -\sum_{q \leq t} g_q \x_q$,
as well as the covariance matrix $\S_t$. Furthermore, the algorithm also keeps track
of a quantity $\Gamma_t \geq 0$, recursively defined as:
\[ 
  \Gamma_0 = 0, \qquad \Gamma_t = \Gamma_{t-1} + g^2_t \x_t^\top \S_t^{\dagger} \x_t.
\]
At the beginning of trial $t$,
after observing $\x_t$ (and updating $\S_t$), the algorithm predicts with weight vector $\w_t$, such that:
\begin{algorithm2e}[t]
\label{alg:two}
\DontPrintSemicolon
\SetAlgoNoEnd
\SetKwInOut{Parameter}{Parameter}
\SetKwInOut{Initialization}{Initialization}
\Parameter{$\alpha > \frac{9}{8}=1.125$.}
\Initialization{$\S \leftarrow \boldsymbol{0}, \; \h \leftarrow \boldsymbol{0}, \Gamma \leftarrow 0$}
 \For{$t=1,\ldots,T$}{
   Receive $\x_t \in \mathbb{R}^d$\;
   Update $\S \leftarrow \S + \x_t \x_t^\top$\;
   $\eta \leftarrow \frac{1}{\alpha} \exp\Big(\frac{1}{2\alpha}\big(\h^\top \S^{\dagger} \h - \Gamma \big)\Big)$ \;
   $\w \leftarrow \eta \S^{\dagger} \h$\;
   Predict with $\hy_t = \w^\top \x_t$\;
   Receive $y_t$ and suffer loss $\ell(y_t,\hy_t)$\;
   Compute $g_t = \partial_{\hy_t} \ell(y_t,\hy_t)$\;
   Update $\h \leftarrow \h - g_t \x_t$\;
   Update $\Gamma \leftarrow \Gamma + g_t^2 \x_t^\top \S^{\dagger} \x_t$\; 
 }
\caption{Scale invariant algorithm}
\end{algorithm2e}
\begin{equation}
  \w_t =  \eta_t \S_t^{\dagger} \h_{t-1}, \qquad 
  \text{where~~} \eta_t = \frac{1}{\alpha} \exp\Big(\frac{1}{2\alpha}\big(\h_{t-1}^\top \S_t^{\dagger} \h_{t-1} - \Gamma_{t-1} \big)\Big).
  \label{eq:weight_update_full}
\end{equation}
This choice of the update leads to the invariance of the algorithm's predictions under transformations of the form $\x_t \mapsto \A \x_t$, $t=1,\ldots,T$, for any invertible matrix $\A$ (shown in Appendix \ref{sec:scale_invariance}).
The algorithm is a second-order method, and is reminiscent of the Online Newton algorithm \citep{logarithmic-regret,Luo_etal2016}. Our algorithm, however, adaptively chooses step size $\eta_t$  (``learning rate'') in each trial. Moreover, no projections are performed, which let us reduce the runtime of the algorithm to $O(d^2)$ per trial (an efficient implementation is discussed at the end of this section). The pseudo-code is presented as Algorithm \ref{alg:two}. 

We now bound the regret of the algorithm. Define the potential function as:
\[
  \psi_t(\h,\S) = 
  \exp\Big(\frac{1}{2 \alpha} \h^\top \S^{\dagger} \h - \Gamma_t\Big)
\]
We have the following result:
\begin{lemma}
\label{lem:full_main_lemma}
In each trial $t=1,\ldots,T$, Algorithm \ref{alg:two} satisfies:
  \[
    g_t \x_t^\top \w_t + \psi_t(\h_t,\S_t) ~\leq~ \psi_{t-1}(\h_{t-1},\S_{t-1}).
  \]
\end{lemma}
The proof is given in Appendix \ref{sec:full_main_lemma}.
The choice of $\w_t$ in Algorithm \ref{alg:two} can be motivated as the one that guarantees bounding the loss of the algorithm by the drop of the potential function (note, however, the as in the coordinate-wise case, the weight vector is not equal to the negative gradient of the potential).
Comparing to Lemma \ref{lem:coordinatewise_main_lemma}, there is no overhead on the right-hand side; however, the overhead is actually hidden in the definition of $\psi_t$ in quantity $\Gamma_t$.

Applying Lemma \ref{lem:full_main_lemma} to each $t=1,\ldots,T$ and summing over trials gives:
\[
  \sum_t g_t \x_t^\top \w_t - \psi_T(\h_T,\S_T) \leq \psi_0(\h_0,\S_0) = 1.
\]
Identifying the left-hand side of the above with the right-hand side
of (\ref{eq:full_penalized_regret_dual}) for $\beta= e^{\frac{\Gamma_T}{2\alpha}}$, we obtain the following bound on the regret:
\begin{align*}
  \regret_T(\u) &\leq \uS \sqrt{\alpha \log\Big(1+ \alpha \uS^2 e^{\frac{\Gamma_T}{\alpha}}\Big)} ~+~ 1 \\
                           &\leq \uS \sqrt{\alpha \log\big(1+ \alpha \uS^2\big) +
                           \log(\alpha) \Gamma_T} ~+~ 1,
\end{align*}
where we used $\log(1+ab) \leq \log(a + ab) = \log a + \log(1+b)$ for $a \geq 1$, applied to $a=e^{\Gamma_T/\alpha} \geq 1$.
Thus, the algorithm achieves an essentially optimal (up to logarithmic factor) scale-invariant bound expressed in terms $\|\u\|_{\S_T}$, with an additional overhead hidden in $\Gamma_T$.

How large can $\Gamma_T$ be? 
By the definition, $\Gamma_T = \sum_t g_t^2 \x_t^\top \S_{t}^{\dagger} \x_t$; as $g_t^2 \x_t^\top \S_{t}^{\dagger} \x_t \leq g_t^2 \leq 1$, $\Gamma_T$ is at most $T$ in the worst case, and the bound becomes
$\tilde{O}(\|\u\|_{\S_T}\sqrt{T})$ (logarithmic factors dropped), which is what
we expected given the negative result in Theorem \ref{thm:gramian_lower_bound}.
However, $\Gamma_T$ can be much smaller in most practical cases as it can be shown to grow only \emph{logarithmically} with the size of the instances \citep[][notation translated to our setup]{Luo_etal2016}:
\begin{lemma}[\citealp{Luo_etal2016}, Theorem 4]
Let $\lambda^*$ be the minimum among the smallest nonzero eigenvalues of
$\S_t$ ($t=1,\ldots, T$) and $r$ be the rank of $\S_T$. We have:
\[
  \sum_{t=1}^T \x_t^\top \S_t^{\dagger} \x_t
  ~\leq~ r + \frac{(1+r)r}{2} \log \bigg(1 + \frac{2 \sum_{t=1}^T \|\x_t\|^2}{(1+r)r \lambda^*} \bigg).
\]
\end{lemma}
Combining the above results, we thus get:
\begin{theorem}
  For any comparator $\u$ and any sequence of outcomes $\{(\x_t,y_t)\}_{t=1}^T$,
  Algorithm \ref{alg:two} satisfies:
\[
  \regret_T(\u) ~\leq~ \uS \sqrt{\alpha \log\big(1+ \alpha \uS^2\big) 
  + \log(\alpha) \Gamma_T} ~+~ 1,
\]
where:
\[
\Gamma_T ~=~ \sum_{t=1}^T g_t \x_t^\top \S_t^{\dagger} \x_t
  ~\leq~ r + \frac{(1+r)r}{2} \log \bigg(1 + \frac{2 \sum_{t=1}^T \|\x_t\|^2}{(1+r)r \lambda^*} \bigg).
\]
with $\lambda^*$ being the minimum among the smallest nonzero eigenvalues of
$\S_t$ ($t=1,\ldots, T$) and $r$ being the rank of $\S_T$.
\label{thm:full}
\end{theorem}

We finally note that the dependence on the dimension $d$ in the bound 
(through the dependence on the rank $r$ in $\Gamma_T$) cannot be eliminated,
as \citet[Theorem 1]{Luo_etal2016} show that in a setting in which the predictions
of $\u$ are constrained to be at most $C$, any algorithm will suffer
the regret at least $\Omega(C\sqrt{dT})$.\footnote{We can, however, improve the dependence on $d$ to $O(\sqrt{d})$ by modifying the algorithm to play with $\tilde{\S}_t = \epsilon \I + \S_t$ for $\epsilon > 0$, and apply the bound on $\sum_t \x_t^\top \tilde{\S}_t \x_t$ from \citet{PLGbook}, Theorem 11.7. This would, however, come at the price of losing the scale invariance of the algorithm.}

\paragraph{Efficient implementation.}
The dominating cost in Algorithm \ref{alg:two} is the computation of pseudoinverse
$\S_t^{\dagger}$ in each trial after performing the update $\S_{t} = \S_{t-1} + \x_t \x_t^\top$, which can be $O(d^3)$. However, we can improve the computational cost per trial to $O(d^2)$ by noticing that $\S_t$ is never used by the algorithm, so it suffices to store and directly update $\S_t^{\dagger}$ using a rank-one update procedure in the spirit Sherman-Morrison formula, which takes $O(d^2$). The procedure is highlighted in the proof of Lemma \ref{lem:full_main_lemma} in Appendix \ref{sec:full_main_lemma}.

\section{Conclusions}

We considered unconstrained online convex optimization, exploiting a natural scale invariance symmetry: the predictions of the optimal comparator (weight vector) are invariant under any linear transformation of the instances (input vectors). Thus, the scale of the weight vector is only relative to the scale of the instances, and we aimed at designing online algorithms which also enjoy this property, i.e. are scale-invariant. We first considered the case of coordinate-wise invariance, in which the individual coordinates (features) can be arbitrarily rescaled. We gave an algorithm, which achieves essentially (up to logarithmic factor) optimal regret bound in this setup (expressed by means of a coordinate-wise scale-invariant norm of the comparator). We then moved to a general (full) invariance with respect to arbitrary linear transformations. We first gave a negative result, showing that no algorithm can achieve a meaningful bound in terms of scale-invariant norm of the comparator in the worst case. Next, we complimented this result with a positive one, providing an algorithm which ``almost'' achieve the desired bound, incurring only a logarithmic overhead in terms of the norm of the instances.

In the future research, we plan to test the introduced algorithms in the computational experiments to verify how their performance relate to the existing online methods from the past work \citep{gd,Ross_etal2013UAI,Orabona_etal2015,Luo_etal2016}.

\acks{%
We thank the anonymous reviewers for suggestions which improved the quality of our work. The author acknowledges support from the Polish National Science Centre (grant no. 2016/22/E/ST6/00299).}

\bibliography{kotlowski}

\begin{thebibliography}{21}
\providecommand{\natexlab}[1]{#1}
\providecommand{\url}[1]{\texttt{#1}}
\expandafter\ifx\csname urlstyle\endcsname\relax
  \providecommand{\doi}[1]{doi: #1}\else
  \providecommand{\doi}{doi: \begingroup \urlstyle{rm}\Url}\fi

\bibitem[Boyd and Vandenberghe(2004)]{BoydVandenberghe04}
S.~Boyd and L.~Vandenberghe.
\newblock \emph{Convex Optimization}.
\newblock Cambridge University Press, 2004.

\bibitem[Campbell and Meyer(2009)]{CambellMayer2009}
Stephen~L. Campbell and Carl~D. Meyer.
\newblock \emph{Generalized Inverses of Linear Transformations}.
\newblock {SIAM}, 2009.

\bibitem[Cesa-Bianchi and Lugosi(2006)]{PLGbook}
Nicol\`o Cesa-Bianchi and G\'abor Lugosi.
\newblock \emph{Prediction, learning, and games}.
\newblock Cambridge University Press, 2006.

\bibitem[Chen and Ji(2011)]{ChenLi2011}
Xuzhou Chen and Jun Ji.
\newblock Computing the {M}oore-{P}enrose inverse of a matrix through symmetric
  rank-one updates.
\newblock \emph{American Journal of Computational Mathematics}, 1\penalty0
  (3):\penalty0 147--151, 2011.

\bibitem[Duchi et~al.(2011)Duchi, Hazan, and Singer]{adagrad}
John~C. Duchi, Elad Hazan, and Yoram Singer.
\newblock Adaptive subgradient methods for online learning and stochastic
  optimization.
\newblock \emph{Journal of Machine Learning Research}, 12:\penalty0 2121--2159,
  2011.

\bibitem[Hazan(2015)]{Hazan_OCO}
Elad Hazan.
\newblock Introduction to online convex optimization.
\newblock \emph{Foundations and Trends in Optimization}, 2\penalty0
  (3--4):\penalty0 157--325, 2015.

\bibitem[Hazan et~al.(2007)Hazan, Agarwal, and Kale]{logarithmic-regret}
Elad Hazan, Amit Agarwal, and Satyen Kale.
\newblock Logarithmic regret algorithms for online convex optimization.
\newblock \emph{Machine Learning}, 69\penalty0 (2-3):\penalty0 169--192, 2007.

\bibitem[Kivinen and Warmuth(1997)]{expgrad}
Jyrki Kivinen and Manfred~K. Warmuth.
\newblock Additive versus {E}xponentiated {G}radient updates for linear
  prediction.
\newblock \emph{Information and Computation}, 132\penalty0 (1):\penalty0 1--64,
  1997.

\bibitem[Luo et~al.(2016)Luo, Agarwal, Cesa-Bianchi, and
  Langford]{Luo_etal2016}
Haipeng Luo, Alekh Agarwal, Nicol{\'o} Cesa-Bianchi, and John Langford.
\newblock Efficient second order online learning by sketching.
\newblock In \emph{Advances in Neural Information Processing Systems ({NIPS})
  29}, 2016.

\bibitem[McMahan and Abernethy(2013)]{McMahanAbernethy2013}
H.~Brendan McMahan and Jacob Abernethy.
\newblock Minimax optimal algorithms for unconstrained linear optimization.
\newblock In \emph{Advances in Neural Information Processing Systems ({NIPS})
  26}, pages 2724--2732, 2013.

\bibitem[McMahan and Orabona(2014)]{McMahanOrabona2014}
H.~Brendan McMahan and Francesco Orabona.
\newblock Unconstrained online linear learning in {H}ilbert spaces: {M}inimax
  algorithms and normal approximation.
\newblock In \emph{Proc. of the 27th Conference on Learning Theory ({COLT})},
  pages 1020--1039, 2014.

\bibitem[McMahan and Streeter(2010)]{McMahanStreeter2010}
H.~Brendan McMahan and Matthew~J. Streeter.
\newblock Adaptive bound optimization for online convex optimization.
\newblock In \emph{Conference on Learning Theory ({COLT})}, pages 244--256,
  2010.

\bibitem[Orabona(2013)]{Orabona2013}
Francesco Orabona.
\newblock Dimension-free exponentiated gradient.
\newblock In \emph{Advances in Neural Information Processing Systems ({NIPS})
  26}, pages 1806--1814, 2013.

\bibitem[Orabona(2014)]{Orabona2014}
Francesco Orabona.
\newblock Simultaneous model selection and optimization through parameter-free
  stochastic learning.
\newblock In \emph{Advances in Neural Information Processing Systems ({NIPS})
  27}, pages 1116--1124, 2014.

\bibitem[Orabona and P{\'a}l(2015)]{OrabonaPal2015ALT}
Francesco Orabona and D{\'a}vid P{\'a}l.
\newblock Scale-free algorithms for online linear optimization.
\newblock In \emph{Algorithmic Learning Theory ({ALT})}, pages 287--301, 2015.

\bibitem[Orabona and P{\'a}l(2016)]{OrabonaPal2016NIPS}
Francesco Orabona and D{\'a}vid P{\'a}l.
\newblock Coin betting and parameter-free online learning.
\newblock In \emph{Neural Information Processing Systems ({NIPS})}, 2016.

\bibitem[Orabona et~al.(2015)Orabona, Crammer, and
  Cesa-Bianchi]{Orabona_etal2015}
Francesco Orabona, Koby Crammer, and Nicol{\`o} Cesa-Bianchi.
\newblock A generalized online mirror descent with applications to
  classification and regression.
\newblock \emph{Machine Learning}, 99\penalty0 (3):\penalty0 411--435, 2015.

\bibitem[Ross et~al.(2013)Ross, Mineiro, and Langford]{Ross_etal2013UAI}
Stephane Ross, Paul Mineiro, and John Langford.
\newblock Normalized online learning.
\newblock In \emph{Proc. of the 29th Conference on Uncertainty in Artificial
  Intelligence ({UAI})}, pages 537--545, 2013.

\bibitem[Shalev-Shwartz(2011)]{ShaiBook}
Shai Shalev-Shwartz.
\newblock Online learning and online convex optimization.
\newblock \emph{Foundations and Trends in Machine Learning}, 4\penalty0
  (2):\penalty0 107--194, 2011.

\bibitem[Streeter and McMahan(2012)]{StreeterMcMahan2012}
Matthew Streeter and H.~Brendan McMahan.
\newblock No-regret algorithms for unconstrained online convex optimization.
\newblock In \emph{Advances in Neural Information Processing Systems ({NIPS})
  25}, pages 2402--2410, 2012.

\bibitem[Zinkevich(2003)]{gd}
Martin Zinkevich.
\newblock Online convex programming and generalized infinitesimal gradient
  ascent.
\newblock In \emph{International Conference on Machine Learning ({ICML})},
  pages 928--936, 2003.

\end{thebibliography}

\appendix

\section{Proof of Lemma \ref{lem:coordinatewise_main_lemma}}
\label{sec:coordinatewise_main_lemma}

Let $t_0$ be the first trial in which $s_{t_0,i} > 0$. This means that in trials
$t=1,\ldots,t_0-1$, $x_{t,i}=0$ and hence $h_{t,i}=0$, and the lemma is trivially satisfied, as the left-hand side is zero. At trial $t_0$, the algorithm still predicts with $w_{t,i}=0$,
and Lemma \ref{lem:coordinatewise_main_lemma} boils down to showing that
\[
  \frac{1}{t_0 d} \exp\Big(\frac{h_{t_0,i}^2}{2 \alpha s_{t_0,i}^2}\Big) \leq \frac{\kappa(\alpha)}{t_0 d},
\]
where $\kappa(\alpha) = e^{\frac{1}{2(\alpha - \alpha_0)}}$. Since $h_{t_0,i} = -g_{t_0} x_{t_0,i}$ and $s^2_{t_0,i} = x^2_{t_0,i}$, we have $\frac{h_{t_0,i}^2}{s_{t_0,i}^2} \leq 1$, and thus the left-hand side is bounded by:
\[
  \frac{1}{t_0 d} e^{\frac{1}{2\alpha}} \leq \frac{1}{t_0 d} e^{\frac{1}{2(\alpha-\alpha_0)}}  =\frac{\kappa(\alpha)}{t_0 d}.
\]
Thus, we have shown the lemma for trials $t=1,\ldots,t_0$. We can now assume that $s_{t-1,i} > 0$ and prove the lemma for the remaining trials $t > t_0$.
By using the definition of $w_{t,i}$ from \eqref{eq:weights_coordinatewise}), we need to show:
\begin{equation}
  \frac{g_t x_{t,i} h_{t-1,i}}{s_{t,i}^2 \alpha t d} \exp\Big(\frac{h_{t-1,i}^2 + x_{t,i}^2}{2 \alpha s_{t,i}^2}\Big) + \frac{1}{t d} \exp\Big(\frac{h_{t,i}^2}{2\alpha s_{t,i}^2}\Big) \leq \frac{1}{(t-1)d} \exp\Big(\frac{h_{t-1,i}^2}{2\alpha s_{t-1,i}^2}\Big) + \frac{\kappa(\alpha)}{td},
  \label{eq:to_show_coordinatewise}
\end{equation}
where we remind that $h_{t,i} = h_{t-1,i} - g_t x_{t,i}$ and $s^2_{t,i} = s^2_{t-1,i} + x_{t,i}^2$.
  First note that the left-hand side is convex in $g_t$, and hence it is maximized for $g_t \in \{-1,1\}$. As the right-hand side does not depend on $g_t$, it suffices to show that the inequality holds for $g_t \in \{-1,1\}$. Furthermore, as the inequality depends on
  $g_t$ only through the product $g_t x_{t,i}$, we assume without loss of generality
  that $x_{t,i} \geq 0$ (the sign can always be incorporated to $g_t$).
We now simplify the notation. Define:
\[
  v = \frac{g_t h_{t-1,i}}{s_{t-1,i}}, \qquad 
  \gamma^2 = \frac{s_{t-1,i}^2}{s^2_{t,i}}.
\]
Note that by the definition $\gamma \in (0,1]$ as $x_{t,i}$ is unconstrained. In this notation, we have:
\begin{align*}
  \frac{x_{t,i}}{s_{t,i}} &~=~ \sqrt{\frac{x^2_{t,i}}{s^2_{t,i}}} =
  \sqrt{1 - \frac{s_{t-1,i}^2}{s^2_{t,i}}} = \sqrt{1 - \gamma^2}, \\
  \frac{h_{t-1,i}^2}{s_{t,i}^2} &~=~
  \frac{h_{t-1,i}^2}{s_{t-1,i}^2} \frac{s_{t-1,i}^2}{s_{t,i}^2}
  = v^2 \gamma^2 \\
  \frac{h_{t,i}^2}{s_{t,i}^2} &~=~
\frac{h_{t-1,i}^2}{s_{t,i}^2}
- 2 \frac{h_{t-1,i} g_t x_{t,i}}{s_{t,i}^2}
+ \frac{x_{t,i}^2}{s_{t,i}^2}
~=~ v^2 \gamma^2 - 2 v \gamma \sqrt{1-\gamma^2} + (1-\gamma^2),
\end{align*}
where we used $g_t^2 = 1$. Using the new notation in \eqref{eq:to_show_coordinatewise}, multiplying both sides by $td$ we equivalently get: 
\begin{equation}
  \frac{v \gamma \sqrt{1-\gamma^2}}{\alpha} e^{\frac{v^2\gamma^2 +1-\gamma^2}{2\alpha}} + e^{\frac{v^2 \gamma^2 - 2v\gamma \sqrt{1-\gamma^2} + 1 - \gamma^2}{2 \alpha}}
  \leq \frac{t}{t-1} e^{\frac{v^2}{2\alpha}} + \kappa(\alpha).
  \label{eq:to_show_coordinatewise_simplied}
\end{equation}
Let us denote the left-hand side of \eqref{eq:to_show_coordinatewise_simplied} as
$A$. We have:
\[
  A = e^{\frac{v^2\gamma^2 +1-\gamma^2}{2\alpha}}\left( \frac{v \gamma \sqrt{1-\gamma^2}}{\alpha} + e^{\frac{-v\gamma \sqrt{1-\gamma^2}}{ \alpha}} \right)
\]
We need the following result, which is proved in Appendix \ref{sec:inequality_exp_squared_exp}:
\begin{lemma}
  Let $\alpha_0 = \frac{9}{8}$. For all $x \in \mathbb{R}$ it holds:
  \[ 
  x + e^{-x} \leq e^{\frac{x^2}{2}\alpha_0}.
 \]
  \label{lem:inequality_exp_squared_exp}
\end{lemma}
Using Lemma \ref{lem:inequality_exp_squared_exp} with $x=\frac{v\gamma\sqrt{1-\gamma^2}}{\alpha}$, we bound:
\[
  A \leq \exp\Big\{\frac{1}{2\alpha} \big(\underbrace{v^2 \gamma^2 + 1 - \gamma^2 + a v^2 \gamma^2 (1-\gamma^2)}_{=B(\gamma^2)} \big) \Big\},
\]
where $a = \frac{\alpha_0}{\alpha} < 1$. Note that $B(\gamma^2)$ is a concave quadratic function of $\gamma^2$, and its (unconstrained) maximizer is given by:
$\gamma^{*2} = \frac{1}{2} + \frac{v^2-1}{2av^2}$. However, since
the allowed values of $\gamma^2$ are $(0,1]$,
the maximizer within the \emph{closure} of this range is:
\begin{itemize}
  \item $\gamma^{*2}=0$ for $v^2 \leq \frac{1}{1+a}$, which gives $B(\gamma^{*2})=1$,
  \item $\gamma^{*2} = 1$ for $v^2 \geq \frac{1}{1-a}$, which gives
    $B(\gamma^{*2})=v^2$,
  \item $\gamma^{*2} = \frac{1}{2} + \frac{v^2-1}{2av^2}$ for
    $v^2 \in \big[ \frac{1}{1+a}, \frac{1}{1-a} \big]$,
    which gives
    $B(\gamma^{*2}) = \frac{a-1}{2a} + v^2 \frac{(1+a)^2}{4a} + \frac{1}{4av^2}$.
    Since this function is monotonically increasing in $v^2$ for $v^2 \geq \frac{1}{1+a}$, we will upper bound it by setting $v^2=\frac{1}{1-a}$, which gives
    $B(\gamma^*) \leq \frac{1}{1-a}$.
\end{itemize}
We thus jointly upper bound $B(\gamma) \leq \max\{v^2,\frac{1}{1-a}\}$, which results
in:
\[
  A \leq \max \Big\{e^{\frac{v^2}{2\alpha}}, e^{\frac{1}{2\alpha(1-a)}}\Big\}
  \leq e^{\frac{v^2}{2\alpha}} + e^{\frac{1}{2\alpha(1-a)}}
  \leq \frac{t}{t-1}e^{\frac{v^2}{2\alpha}} + \kappa(\alpha),
\]
which verifies \eqref{eq:to_show_coordinatewise_simplied} and finishes the proof.

\section{Proof of Lemma \ref{lem:conjugate}}
\label{appendix:conjugate}
Without loss of generality assume $\A$ has $k \geq 1$ strictly positive eigenvalues
(the remaining eigenvalues being zero), as otherwise (for $k=0$)
$\mathrm{range}(\A) = \{\boldsymbol{0}\}$ and there is nothing to show.
Let $\A = \V \SSigma \V^\top$ for $\V \in \mathbb{R}^{d \times k}$,
$\SSigma = \mathrm{diag}(\lambda_1,\ldots,\lambda_k)$, be the `thin' eigenvalue
decomposition of $\A$ (i.e., the eigendecomposition without explicit appearance
of eigenvectors with zero eigenvalues). Since $\y \in \mathrm{range}(\A)$,
there exists $\widetilde{\y} \in \mathbb{R}^d$, such that
$\y = \A \widetilde{\y}$, and therefore:
\[
  \y ~=~ \V \SSigma \V^\top \widetilde{\y} ~=~
  \V \SSigma^{1/2} \z, \qquad \text{where~} \z = \SSigma^{1/2} \V^\top \widetilde{\y} \in
  \mathbb{R}^k.
\]
We have:
\begin{align*}
\sup_{\u \in \mathbb{R}^d} \bigg\{ \y^\top \u - f\Big(\sqrt{\u^\top \A \u}\Big)
 \bigg\} 
 &~=~ 
\sup_{\u \in \mathbb{R}^d} \bigg\{ \z^\top (\SSigma^{1/2} \V^\top \u) - 
f\Big(\big \|\SSigma^{1/2} \V^\top \u \big \|\Big)
 \bigg\}  \\
 &~=~ \sup_{\u \in \mathbb{R}^k} \Big\{ \z^\top \u - 
f(\|\u \|) \Big\},
\end{align*}
where we reparametrized $\SSigma^{1/2} \V^\top \u$ as $\u \in \mathbb{R}^k$.
Now, note that keeping the norm $\|\u\|$ fixed, the supremum
is achieved by $\u$ in the direction of $\z$; therefore, without loss
of generality assume $\u = \beta \frac{\z}{\|\z\|}$ for some $\beta \geq 0$.
This means that:
\[
\sup_{\u \in \mathbb{R}^d} \big\{ \z^\top \u -
 f(\|\u\|)
 \big\} = \sup_{\beta \geq 0} \big\{
    \beta \|\z\| - f(\beta) \big\}
 = f^*(\|\z\|).
\]
Since $\A^{\dagger} = \V \SSigma^{-1} \V^\top$,
\[
  \|\y\|^2_{\A^{\dagger}} = \y^\top \A^{\dagger} \y
  = \y^\top \V \SSigma^{-1} \V^\top \y
  = \| \SSigma^{-1/2} \V^\top \y \|^2
  = \| \z\|^2,
\]
which finishes the proof of the first part of the lemma.

For the second part,
we only need $\h_T \in \mathrm{range}(\S_T)$, a well-known fact, which
we show below for completeness.
Let $\v$ be any eigenvector of $\S_T$ associated with zero eigenvalue,
so that $\v^\top \S_T \v = 0$. Using the definition
$\S_T = \sum_{t=1}^T \x_t \x_t^\top$, we have 
$\v^\top \S_T \v = \sum_{t=1}^T (\x_t^\top \v)^2 = 0$,
which implies $\x_t^\top \v = 0$ for all $t$. But since 
$\h_T = -\sum_{t=1}^T g_t \x_t$, this also means $\h_T^\top \v = 0$.
Let $\v_1,\ldots,\v_k$ be the eigenvectors
of $\S_T$ associated with all non-zero eigenvalues $\lambda_1,\ldots,\lambda_k$.
By the previous argument, $\h_T \in \mathrm{span}\left\{\v_1,\ldots,\v_k\right\}$, i.e.
$\h_T = \sum_{i=1}^k \alpha_i \v_i$. Choosing a vector $\z = \sum_{i=1}^k \frac{\alpha_i}{\lambda_i} \v_i$ reveals that:
\[
  \S_T \z = \sum_{i=1}^k \lambda_i \v_i \v_i^\top \z
  = \sum_{i=1}^k \lambda_i \frac{\alpha_i}{\lambda_i} \v_i
  = \sum_{i=1}^k \alpha_i \v_i = \h_T,
\]
which shows that $\h_T \in \mathrm{range}(\S_T)$.
This finishes the proof. 

\section{Proof of Theorem \ref{thm:gramian_lower_bound}}

We will show that for any algorithm, there exists a sequence of outcomes $\{(\x_t,g_t)\}_{t=1}^T$, such that the loss of the algorithm is nonnegative on this sequence, while
$\h_T^\top \S_T^{\dagger} \h_T = \frac{T}{2}$. Then, we can choose the comparator as
$\u = \beta \sqrt{\frac{2}{T}} \, \S_T^{\dagger} \h_T$, which invariant norm is 
equal to:
\[
  \|\u\|_{\S_T} = \sqrt{\u^\top \S_T \u}
  =  \beta \sqrt{\frac{2}{T}} \sqrt{\h_T^\top \S_T^{\dagger} \S_T \S_T^{\dagger} \h_T}
  =  \beta \sqrt{\frac{2}{T}} \sqrt{\h_T^\top \S_T^{\dagger} \h_T}
  =  \beta \sqrt{\frac{2}{T}} \sqrt{\frac{T}{2}} = \beta,
\]
and the regret becomes:
\begin{align*}
  \regret_T(\u) 
  &= \underbrace{\sum_{t=1}^T g_t \x_t^\top \w_t}_{\textrm{nonnegative}}
  + \h_T^\top \u 
\geq \h_T^\top \u = \beta \sqrt{\frac{2}{T}} \h_T^\top \S_T^{\dagger} \h_T 
= \beta \sqrt{\frac{T}{2}}
=  \|\u\|_{\S_T} \sqrt{\frac{T}{2}}.
 \end{align*}
Thus, to finish the proof, it suffices to find a sequence with the claimed
properties. The sequence goes as follows. In each trial $t$, the adversary chooses
the sign of $g_t$ to match the sign of $\widehat{y}_t = \x_t^\top \w_t$, 
so that $g_t \x_t^\top \w_t \geq 0$, i.e. the loss of the algorithm is
nonnegative in each trial. Moreover, $\x_t$ and $g_t$ are chosen so that
$\h_t^\top \S_t^{\dagger} \h_t = \frac{t}{2}$ for all $t$,
and hence $\h_T^\top \S_T^{\dagger} \h_T = \frac{T}{2}$.
In the first $d$ trials, the adversary chooses
$\x_t = \e_t$ and $g_t \in \{-\frac{1}{\sqrt{2}},\frac{1}{\sqrt{2}}\}$. 
In this case, for any $t=1,\ldots,d$:

$$\h_t = (g_1,\ldots,g_t,0,\ldots,0), \qquad \S_t = \S_t^{\dagger} 
= \mathrm{diag}(\underbrace{1,\ldots,1}_{t},0\ldots,0),$$
and hence
$\h_t^\top \S_t^{\dagger} \h_t = \sum_{i=1}^t g_i^2 = \frac{t}{2}$. 
Note that $\G_t$ is invertible for all $t \geq d$, i.e. $\S_t^{\dagger} = \S_t^{-1}$
for $t \geq d$.
For the remaining trials $t=d+1,\ldots,T$ the adversary chooses $g_t \in \{-1,1\}$,
and $\x_t$ in any 
direction for which $\h^\top_{t-1} \S^{-1}_{t-1} \x_t = 0$, with $\|\x_t\|$ chosen
in such a way that $\x_t^\top \S_{t-1}^{-1} \x_t = 1$ (which is always possible as $d \geq 2$). Using Sherman-Morrison 
inversion formula for $\S_t = \S_{t-1} + \x_t \x_t^\top$,
\begin{align*}
  \h_t^\top \S_t^{\dagger} \h_t^\top
  &= \h_t^\top \S_{t-1}^{-1} \h_t
  - \frac{\big(\h_t^\top\S_{t-1}^{-1}\x_t\big)^2}{1 + \x_t^\top \S_{t-1}^{-1} \x_t}
  = \h_{t-1}^\top \S_{t-1}^{-1} \h_{t-1}
    + \x_t^\top \S_{t-1}^{-1} \x_t 
    - \frac{\big(\x_t^\top\S_{t-1}^{-1}\x_t\big)^2}
    {1 + \x_t^\top \S_{t-1}^{-1} \x_t} \\
  &= \h_{t-1}^\top \S_{t-1}^{-1} \h_{t-1}
  + \frac{\x_t^\top \S_{t-1}^{-1} \x_t}{1 + \x_t^\top \S_{t-1}^{-1} \x_t}
  = \h_{t-1}^\top \S_{t-1}^{\dagger} \h_{t-1} + \frac{1}{2},
\end{align*} 
where in the second equality we used $\h_{t-1}^\top \S^{-1}_{t-1} \x_t = 0$ and
in the last equality we used $\x_t^\top \S_{t-1}^{-1} \x_t = 1$. Thus,
$\h_t^\top \S_t^{\dagger} \h_t = \frac{t}{2}$ for all $t$.

\section{Scale invariance of Algorithm \ref{alg:two}}
\label{sec:scale_invariance}
We need to show that under linear transformation $\x_t \mapsto \A \x_t$, $t=1,\ldots,T$, for any invertible $\A$, the predictions of the algorithm, $\{\hy_t\}_{t=1}^T$, do not change. We remind that the predictions are given by:
\[
  \hy_t =  \eta_t \x_t^\top \S_t^{\dagger} \h_{t-1}, \quad 
  \text{where~~} \eta_t = \frac{1}{\alpha} \exp\Big(\frac{1}{2\alpha}\big(\h_{t-1}^\top \S_t^{\dagger} \h_{t-1} - \Gamma_{t-1} \big)\Big),
  \quad
  \Gamma_t = \sum_{q \leq t} g_q^2 \x_q^\top \S_t^{\dagger} \x_q.
\]
We proceed by induction on $t$. For $t=1$, $\hy_1 = 0$, which is trivially invariant under linear transformation of the instances. Now, assume inductively that $\{\hy_q\}_{q=1}^{t-1}$ are invariant, which implies $\{g_q\}_{q=1}^{t-1}$ are also invariant as they only depend on past predictions and past labels. Since $\h_{t-1} = -\sum_{q < t} g_q \x_q$, prediction $\hy_t$ depends on the data only by means of $\{g_q\}_{q=1}^{t-1}$  and quantities $\x_i^\top \S_t^{\dagger} \x_j$ for $i,j \leq t$.
Thus, to show the invariance of $\hy_t$ under linear transformation of the instances, it suffices to show the invariance of $\x_i^\top \S_t^{\dagger} \x_j$, for $i,j \leq t$.
As $\S_t = \sum_{q \leq t} \x_q \x_q^\top$ maps to $\sum_{q \leq t} \A \x_q \x_q^\top \A^\top = \A \S_t \A^\top$ under linear transformation, it thus amounts to show that:
\[
  \x_i^\top \S_t^{\dagger} \x_j ~=~ (\A \x_i)^\top \left(\A \S_t \A^\top \right)^{\dagger} \A \x_j
  ~=~ \x_i^\top \A^\top \left(\A \S_t \A^\top \right)^{\dagger} \A \x_j,
\]
for any $i,j \leq t$ and any invertible $\A$.\footnote{The simplest approach to show the invariance would be to prove $(\A \S_t \A^\top)^{\dagger} = \A^{-\top} \S_t^{\dagger} \A^{-1}$. Unfortunately (and surprisingly to us), while this relation holds for a matrix inverse, it \emph{does not} hold in general for pseudoinverse! In the proof, we need to use a crucial fact that $\x_i$ and $\x_j$ are in the range of $\S_t$.}
Since $\x_i, \x_j \in \mathrm{range}(\S_t)$, we have: 
$\x_i = \S_t^{\dagger} \S_t  \x_i$ and similarly $\x_j = \S_t^{\dagger} \S_t \x_j$. Thus:
\begin{align*}
\x_i^\top \A^\top \left(\A \S_t \A^\top \right)^{\dagger} \A \x_j
&~=~ \x_i^\top \S_t^{\dagger} \S_t \A^\top \left(\A \S_t \A^\top \right)^{\dagger} \A \S_t \S_t^{\dagger} \x_j \\
&~=~ \x_i^\top \S_t^{\dagger} \A^{-1} \left(\A \S_t \A^\top\right) \left(\A \S_t \A^\top \right)^{\dagger} \left( \A \S_t \A^{\top}\right) \A^{-\top} \S_t^{\dagger} \x_j \\
&~=~ \x_i^\top \S_t^{\dagger} \A^{-1} \left(\A \S_t \A^\top\right) \A^{-\top} \S_t^{\dagger} \x_j \\
&~=~ \x_i^\top \S_t^{\dagger} \S_t \S_t^{\dagger} \x_j \\
&~=~ \x_i^\top \S_t^{\dagger} \x_j,
\end{align*}
which was to be shown. 
\section{Proof of Lemma \ref{lem:full_main_lemma}}
\label{sec:full_main_lemma}
By plugging the definition of the algorithm's weight (\ref{eq:weight_update_full})
to the inequality in the lemma,
we need to show:
\begin{equation}
  \frac{g_t \x_t^\top \S_t^{\dagger} \h_{t-1} }{\alpha} e^{\frac{1}{2\alpha}(\h_{t-1}^\top \S_t^{\dagger} \h_{t-1} - \Gamma_{t-1})}
  +e^{\frac{1}{2 \alpha} (\h_t^\top \S^{\dagger}_t \h_t - \Gamma_t)}
  \leq e^{\frac{1}{2 \alpha} (\h_{t-1}^\top \S_{t-1}^{\dagger} \h_{t-1} - \Gamma_{t-1})}.
\label{eq:to_show_full}
\end{equation} 
Using $\h_t = \h_{t-1} - g_t \x_t$, we have:
\begin{align*}
\h_t^\top \S^{\dagger}_t \h_t - \Gamma_t
&= \h_{t-1}^\top \S^{\dagger}_t \h_{t-1} - 2g_t \h_{t-1}^\top \S^{\dagger}_t \x_t + g_t^2 \x_t^\top \S^{\dagger}_t \x_t - \Gamma_t\\
&=\h_{t-1}^\top \S^{\dagger}_t \h_{t-1} - 2g_t \h_{t-1}^\top \S^{\dagger}_t \x_t 
- \Gamma_{t-1},
\end{align*}
from the definition of $\Gamma_t$. Plugging the above into \eqref{eq:to_show_full}
and multiplying both sides by $e^{\Gamma_{t-1}}$, we equivalently need to show:
\begin{equation}
  e^{\frac{1}{2\alpha}\h_{t-1}^\top \S_t^{\dagger} \h_{t-1}}
  \Big(
  \frac{g_t \h_{t-1}^\top \S^{\dagger}_t \x_t}{\alpha}
  + e^{- \frac{g_t \h_{t-1}^\top \S^{\dagger}_t \x_t }{\alpha}}
\Big)
\leq e^{\frac{1}{2 \alpha} \h_{t-1}^\top \S_{t-1}^{\dagger} \h_{t-1}} .
\label{eq:to_show_full_simpler}
\end{equation} 
Denote the left-hand side of \eqref{eq:to_show_full_simpler} by $A$.
Applying $x + e^{-x} \leq e^{\frac{x^2}{2} \alpha_0}$ for $\alpha_0=\frac{9}{8}$ (Lemma \ref{lem:inequality_exp_squared_exp}) with $x = \frac{g_t \h_{t-1}^\top \S^{\dagger}_t \x_t }{\alpha}$ to the left-hand side of \eqref{eq:to_show_full_simpler}
results in the following bound:
\begin{align*}
  A &~\leq~ \exp \Big(\frac{1}{2\alpha} \big(\h_{t-1}^\top \S_t^{\dagger} \h_{t-1}
  + \frac{\alpha_0}{\alpha} \big(g_t \h_{t-1}^\top \S^{\dagger}_t \x_t \big)^2 \big) \Big) \\
  &~\leq~ \exp \Big(\frac{1}{2\alpha} \big(\h_{t-1}^\top \S_t^{\dagger} \h_{t-1}
  + \big(\h_{t-1}^\top \S^{\dagger}_t \x_t\big)^2 \big) \Big), 
\end{align*}
where we used $\alpha_0 \leq \alpha$ and $g_t^2 \leq 1$.

We now express $\S_t^{\dagger}$ in terms of $\S_{t-1}^{\dagger}$, by using an extension of well-known Sherman-Morisson formula to pseudoinverse \citep{CambellMayer2009,ChenLi2011}. To this end define $\x_{\perp} = (\I - \S_{t-1}\S_{t-1}^{\dagger} ) \x_t$
to be the component of $\x_t$ orthogonal to the range of $\S_{t-1}$. Note that
since $\h_{t-1} \in \mathrm{range}(\S_{t-1})$ (see the proof of Lemma \ref{lem:conjugate} in Appendix \ref{appendix:conjugate}), we have $\h_{t-1}^\top \x_{\perp} = 0$.
Moreover, $\x_{\perp}^\top \x_t = \x^\top_t (\I - \S_{t-1}\S_{t-1}^{\dagger} ) \x_t =
\x^\top_t (\I - \S_{t-1}\S_{t-1}^{\dagger} )^2 \x_t = \x_{\perp}^\top \x_{\perp} = \|\x_{\perp}\|^2$, where we used that $\I - \S_{t-1}\S_{t-1}^{\dagger}=(\I - \S_{t-1}\S_{t-1}^{\dagger})^2$ is idempotent as a projection operator.
Let us also define $\beta = 1 + \x_t^\top \S_{t-1}^{\dagger} \x_t$.
Depending on whether $\x_{\perp} = \boldsymbol{0}$ we need to consider two cases \citep{ChenLi2011}:
\begin{enumerate}
  \item Case $\x_{\perp} = \boldsymbol{0}$. In this case, we essentially follow Sherman-Morrison formula:
    \[
      \S_t^{\dagger} = \S_{t-1}^{\dagger} - \frac{1}{\beta} \S_{t-1}^{\dagger} \x_t \x_t^\top \S_{t-1}^{\dagger},
    \]
and we get:
\begin{align*}
  \h_{t-1}^{\top} \S_t^{\dagger} \h_{t-1} &= \h_{t-1}^{\top} \S_{t-1}^{\dagger} \h_{t-1} 
  - \frac{1}{\beta} \big(\h_{t-1}^\top \S_{t-1}^{\dagger} \x_t \big)^2,\\
    \h_{t-1}^\top \S_t^{\dagger} \x_t &=
    \h_{t-1}^\top \S_{t-1}^{\dagger} \x_t 
    - \frac{1}{\beta} \h_{t-1}^\top \S_{t-1}^{\dagger} \x_t 
    \underbrace{\x_t^\top \S_{t-1}^{\dagger} \x_t}_{=\beta-1}  = 
    \frac{1}{\beta} \h_{t-1}^\top \S_{t-1}^{\dagger} \x_t.
\end{align*}
Therefore:
\begin{align*}
  A &~\leq~ \exp \bigg(\frac{1}{2\alpha} \Big(
      \h_{t-1}^\top \S_{t-1}^{\dagger} \h_{t-1} 
  - \frac{1}{\beta} \big(\h_{t-1}^\top \S_{t-1}^{\dagger} \x_t \big)^2
+ \frac{1}{\beta^2} \big(\h_{t-1}^\top \S_{t-1}^{\dagger} \x_t \big)^2\Big)
  \bigg) \\
  &~\leq~ \exp \Big(\frac{1}{2\alpha} 
\h_{t-1}^\top \S_{t-1}^{\dagger} \h_{t-1} \Big),
\end{align*}
where we used $\beta \geq 1$. Thus, \eqref{eq:to_show_full_simpler} follows.
\item Case $\x_{\perp} \neq \boldsymbol{0}$. In this case the update formula
  has a different form:
\[
  \S_t^{\dagger} = \S_{t-1}^{\dagger}
  - \frac{\S_{t-1}^{\dagger}\x_t \x_{\perp}^\top}{\|\x_{\perp}\|^2}
  - \frac{\x_{\perp} \x_t^\top \S_{t-1}^{\dagger}}{\|\x_{\perp}\|^2}
+ \beta \frac{\x_{\perp} \x_{\perp}^\top}{\|\x_{\perp}\|^4},
\]
and we get:
\begin{align*}
  \h_{t-1}^\top \S_t^{\dagger} \h_{t-1} 
  &= \h_{t-1}^\top \S_{t-1}^{\dagger} \h_{t-1},\\
    \h_{t-1}^\top \S_t^{\dagger} \x_t &=
    \h_{t-1}^\top \S_{t-1}^{\dagger} \x_t 
    - \h_{t-1}^\top \S_{t-1}^{\dagger} \x_t \frac{\x_{\perp}^\top \x_t}{\|\x_{\perp}\|^2} = 0,
\end{align*}
where we used $\h^\top \x_{\perp} = 0$ and $\x_{\perp}^\top \x_t = \|\x_{\perp}\|^2$.
Therefore,
\[
  A ~\leq~ \exp \Big(\frac{1}{2\alpha} 
\h_{t-1}^\top \S_{t-1}^{\dagger} \h_{t-1} \Big),
\]
and \eqref{eq:to_show_full_simpler} follows. This finishes the proof.
\end{enumerate}

\section{Proof of Lemma \ref{lem:inequality_exp_squared_exp}}
\label{sec:inequality_exp_squared_exp}
We need to show that for all $x \in \mathbb{R}$ it holds:
  \begin{equation}
   x + e^{-x} \leq e^{\frac{x^2}{2}\alpha_0}.
  \label{eq:inequality_exp_squared_exp}
  \end{equation}  
  where $\alpha_0 = \frac{9}{8}$. First, consider $x > 0$. We have
  $e^{\frac{x^2}{2}\alpha_0} \geq e^{\frac{x^2}{2}} \geq 1 + \frac{x^2}{2}$,
  and $e^{x} \geq 1 + x + \frac{x^2}{2}$, which implies 
  $e^{-x} \leq \frac{1}{1+x+\frac{x^2}{2}}$. Thus, it suffices to show 
$1+\frac{x^2}{2} \geq \frac{1}{1+x+\frac{x^2}{2}} + x$, which amounts to:
\[
  \frac{x^2}{2}-x+1 \geq \frac{1}{1+x+\frac{x^2}{2}}
  \; \iff \;
  \bigg(\frac{x^2}{2}+1-x\bigg)\bigg(\frac{x^2}{2}+1+x\bigg) \geq 1
  \; \iff \;
  \frac{x^4}{4} \geq 0,
\]
which clearly holds. Now, consider $x \leq 0$. Showing (\ref{eq:inequality_exp_squared_exp}) for $x<0$ is equivalent to showing:
\begin{equation}
e^{\frac{x^2}{2}\alpha_0} +x - e^{x} \geq 0,
\label{eq:inequality_exp_squared_exp_minus}
\end{equation}
for $x \geq 0$ (after substituting $x \to -x$). First note that when $x \geq \frac{2}{\alpha_0}$,
$e^{\frac{x^2}{2}\alpha_0} \geq e^{x} \geq e^{x} - x$, so that (\ref{eq:inequality_exp_squared_exp_minus}) holds. Thus, it suffices to check the inequality for $x \in [0,2/\alpha_0] = [0,16/9]$.
When $x \leq 0.34$, we make use of the fact that $\frac{e^x-x-1}{x^2}$ is nondecreasing in $x$, so that $e^x-x \leq 1 + x^2 \frac{e^{0.34}-0.34-1}{0.34^2}
\leq 1 + 0.5619 x^2$, whereas $e^{\frac{x^2}{2}\alpha_0} \geq 1 + \frac{x^2}{2}\alpha_0
\geq 1 + \frac{9}{16} x^2 = 1 + 0.5625 x^2$. Therefore, $e^{\frac{x^2}{2}\alpha_0} - e^x + x \geq 0.0006 x^2 \geq 0$, and we showed (\ref{eq:inequality_exp_squared_exp_minus}) for $x \leq 0.34$.
From now on, the proof becomes very tedious and requires first-order Taylor approximations and the use of convexity at various subintervals of $[0.34,\frac{16}{9}]$ to finally combine the bound.  Thus, the rest of the proof works by splitting the range $[0.34,\frac{16}{9}]$ into
intervals $[u_1,v_1],\ldots,[u_m,v_m]$, where $u_1 = 0.34$, $u_i = v_{i-1}$, and
$v_m = \frac{16}{9}$. For each $i=1,\ldots,m$, for $x \in [u_i,v_i]$, we use convexity of $e^x$ and upper bound it by:
\[
  e^x \leq \frac{x-u_i}{v_i-u_i}e^{v_i} + \frac{v_i - x}{v_i -u_i} e^{u_i}
  = \underbrace{\frac{v_i e^{u_i} - u_i e^{v_i}}{v_i - u_i}}_{=b_i} + x \underbrace{\frac{e^{v_i} - e^{u_i}}{v_i-u_i}}_{=c_i}.
\]
On the other hand, we use convexity of $f(x) = e^{\frac{x^2}{2}\alpha_0}$ to lower bound it
by $f(x) \geq f(u_i) + f'(u_i)(x-u_i)$. Thus, the left-hand side of (\ref{eq:inequality_exp_squared_exp_minus}) is lower bounded by:
\begin{equation}
  f(u_i) -f'(u_i) u_i - b_i + x(f'(u_i)-c_i+1)
  \geq f(u_i) -f'(u_i) u_i - b_i + \min\{v_i(f'(u_i)-c_i+1), u_i(f'(u_i)-c_i+1) \}.
  \label{eq:lower_bound_technical}
\end{equation}
In the table below, we present the numerical values of (\ref{eq:lower_bound_technical}) for a set of chosen intervals:
\begin{center}
  \begin{tabular}{c@{~~~~~}c}
\toprule
interval $[u_i,v_i]$ & lower bound (\ref{eq:lower_bound_technical}) \\
\hline
$[1.24,\,1.78]$ & 0.017 \\ 
$[0.99,\,1.24]$ & 0.003 \\ 
$[0.85,\,0.99]$ & 0.001 \\ 
$[0.76,\,0.85]$ & 0.0007 \\ 
$[0.7,\,0.76]$ & 0.001 \\ 
$[0.65,\,0.7]$ & 0.0008 \\ 
$[0.6,\,0.65]$ & 0.0002 \\ 
$[0.56,\,0.6]$ & 0.0008 \\ 
$[0.52,\,0.56]$ & 0.0005 \\ 
$[0.47,\,0.52]$ & 0.0002 \\ 
$[0.42,\,0.47]$ & 0.0004 \\ 
$[0.37,\,0.42]$ & 0.0005 \\ 
$[0.34,\,0.37]$ & 0.001 \\ 
\bottomrule
\end{tabular}
\end{center}
As all lower bounds are positive, this finishes the proof.

\paragraph{Note:} If we chose $\alpha_0 = 2$, the proof of the lemma would simplify dramatically, as we would only need to separately bound $x \in (-\infty,-1)$, $x \in [-1,0]$, and $x \in (0,\infty)$. However, we opted for the smallest $\alpha_0$, as smaller $\alpha_0$ translates to a smaller achievable constant in the regret bound. Our choice $\alpha_0 = \frac{9}{8}$ was obtained by performing numerical tests, which showed that this value is very close to the smallest $\alpha_0$, for which the inequality still holds.

\end{document}